\documentclass[journal]{IEEEtran}

\ifCLASSINFOpdf
\else
   \usepackage[dvips]{graphicx}
\fi
\usepackage{url}
\usepackage{tabularx}

\usepackage{amsmath,amssymb,bm,amsfonts}
\usepackage{graphicx}
\usepackage{textcomp}
\usepackage{xcolor}
\usepackage{booktabs}
\usepackage{multirow}
\usepackage[normalem]{ulem}
\usepackage{url}
\usepackage{amsthm}
\usepackage{subfigure}
\usepackage{stfloats}
\usepackage{bbding}
\usepackage{diagbox}
\usepackage{soul} % colar
\usepackage{color, xcolor} % 

\usepackage{graphicx}
\usepackage{cite}      % 自己加的
\usepackage{hyperref}  % 自己加的
\usepackage{amsthm,amsmath,amssymb} % 自己加的，花体
\usepackage{mathrsfs}  % 自己加的，花体
\usepackage{booktabs}
\usepackage{bm,balance}
\hypersetup{hidelinks}
\usepackage[normalem]{ulem}
\useunder{\uline}{\ul}{}

\usepackage{tikz}% Make Orcid icon
\definecolor{lime}{HTML}{A6CE39}
\DeclareRobustCommand{\orcidicon}{%
    \begin{tikzpicture}
    \draw[lime, fill=lime] (0,0) 
    circle [radius=0.16] 
    node[white] {{\fontfamily{qag}\selectfont \tiny ID}};    \draw[white, fill=white] (-0.0625,0.095) 
    circle [radius=0.007];    \end{tikzpicture}
    \hspace{-2mm}}
\foreach \x in {A, ..., Z}{%
    \expandafter\xdef\csname orcid\x\endcsname{\noexpand\href{https://orcid.org/\csname orcidauthor\x\endcsname}{\noexpand\orcidicon}}
}

\begin{document}

\title{ Multi-scale Spatial-Temporal Interaction Network for Video Anomaly Detection }
% Object-centric Spatiotemporal residual Network for Video Anomaly Detection
\author{Zhiyuan Ning\orcidA{}, Zhangxun Li\orcidB{}, Zhengliang Guo\orcidC{}, Zile Wang\orcidD{}, Liang Song\orcidE{},~\IEEEmembership{Senior Member,~IEEE}

% \thanks{This paragraph of the first footnote will contain the date on which you submitted your paper for review. It will also contain support information, including sponsor and financial support acknowledgment. For example, ``This work was supported in part by the U.S. Department of Commerce under Grant BS123456.'' }
% \thanks{The next few paragraphs should contain the authors' current affiliations, including current address and e-mail. For example, F. A. Author is with the National Institute of Standards and Technology, Boulder, CO 80305 USA (e-mail: author@boulder.nist.gov).}
\thanks{This study is jointly funded by the Fudan University-Changan Joint Lab on Networked AI Edge Computing, the China Mobile Research Fund under the Chinese Ministry of Education (Grant No. KEH2310029), and the Shanghai Key Research Lab of NSAI. (Corresponding author: Liang Song.)}
\thanks{Zhiyuan Ning, Zhangxun Li, Zhengliang Guo, and Liang Song are with the Academy for Engineering \& Technology, Fudan University, Shanghai 200433, China (e-mails: 22210860109@m.fudan.edu.cn;  songl@fudan.edu.cn).}
\thanks{Zile Wang is with the School of Information Science and Technology, Fudan University, Shanghai 200433, China.}
% \thanks{The authors are with the Academy for Engineering \& Technology, Fudan University, Shanghai 200433, China (e-mail: songl@fudan.edu.cn).}
}

\markboth{Journal of \LaTeX\ Class Files, Vol. 14, No. 8, August 2015}
{Shell \MakeLowercase{\textit{et al.}}: Bare Demo of IEEEtran.cls for IEEE Journals}
\maketitle

\begin{abstract}
Video Anomaly Detection (VAD) is an essential yet challenging task in signal processing. Since certain anomalies cannot be detected by isolated analysis of either temporal or spatial information, the interaction between these two types of data is considered crucial for VAD. However, current dual-stream architectures either confine this integral interaction to the bottleneck of the autoencoder or introduce anomaly-irrelevant background pixels into the interactive process, hindering the accuracy of VAD. To address these deficiencies, we propose a Multi-scale Spatial-Temporal Interaction Network (MSTI-Net) for VAD. First, to prioritize the detection of moving objects in the scene and harmonize the substantial semantic discrepancies between the two types of data, we propose an Attention-based Spatial-Temporal Fusion Module (ASTFM) as a substitute for the conventional direct fusion. Furthermore, we inject multi-ASTFM-based connections that bridge the appearance and motion streams of the dual-stream network, thus fostering multi-scale spatial-temporal interaction. Finally, to bolster the delineation between normal and abnormal activities, our system records the regular information in a memory module. Experimental results on three benchmark datasets validate the effectiveness of our approach, which achieves AUCs of 96.8\%, 87.6\%, and 73.9\% on the UCSD Ped2, CUHK Avenue, and ShanghaiTech datasets, respectively.
\end{abstract}


\IEEEpeerreviewmaketitle

\section{Introduction}  % 标题用\section

\IEEEPARstart{V}{ideo} Anomaly Detection (VAD) is a meaningful task in the signal processing community\cite{cheng2023learning,liu2023distributional,li2022adaptive} as its objective is to automate the detection of unusual events in large-scale video sequences\cite{liu2023generalized,Tang2020,Zhou2019}. However, the low incidence and ambiguous composition of abnormal events make the data collection and analysis time-consuming\cite{zhao2022exploiting}. Therefore, 
a prevalent strategy is to deploy an unsupervised learning model that is solely trained on normal data\cite{zaheer2022generative}, and activities or events identified as outliers by this model are treated as anomalies\cite{le2023attention}. Under this paradigm, the essence of unsupervised VAD lies in modeling the spatial-temporal information of regular events or activities. Beyond the isolated representation of each data type, the cardinal task is to model their correlation\cite{zhao2022lgn}.

In real-life situations, anomalies come in many forms and can be broadly categorized into appearance-only (e.g., blazing fire, knife-wielding thug), motion-only (e.g., fighting, robbing), and appearance-motion joint anomalies (e.g., the car driving towards pedestrians)\cite{liu2023osin}. The detection of appearance-motion joint anomalies poses a particular challenge as it necessitates more than the separate analysis of spatial or temporal data. To illustrate this, consider the scenario of a car colliding with a pedestrian at a normal speed. Individually, the appearance of both the car and the pedestrian is considered normal. Similarly, the motion of each does not reveal any irregular changes. Consequently, the identification of this joint appearance-motion anomaly calls for modeling the consistency between the two data types through spatial-temporal interactions\cite{liu2022learning}.

Previous single-stream networks mainly engage in modeling the spatial aspects of regular events, while overlooking the analysis of temporal anomalies, including reconstruction\cite{Hasan2016} and prediction\cite{ffp,liu2023msn, Mathieu2016}. Moreover, numerous existing methods resort to dual-stream networks to independently learn spatial and temporal patterns of regular events\cite{vu2019robust,yan2018abnormal,xu2017detecting}, yet they 
stumble in considering their interaction. In response, many researchers incorporate spatial-temporal consistency semantics into dual-stream networks\cite{hao2022spatiotemporal,AMMC}. However, these strategies typically confine the spatial-temporal interaction to the autoencoder's bottleneck, restricting their capacity to effectively learn intricate spatial-temporal consistency semantics\cite{cheng2023spatial}.

To address these challenges, we propose a Multi-scale Spatial-Temporal Interaction Network (MSTI-Net) that incorporates spatial-temporal interactions across a multi-scale spatial-temporal receptive field. Traditional interaction methods tend to integrate spatial-temporal data in a simple manner (e.g., concatenation of feature\cite{ffp}). Besides, they neglect that the appearance stream involves abundant background pixels irrelevant to anomalies, resulting in performance degradation. In response, we propose an Attention-based Spatial-Temporal Fusion Module (ASTFM). ASTFM assists dual-stream networks in reconciling significant semantic differences and focuses on moving objects while disregarding background information. Furthermore, we establish multiple ASTFM-based pathways between two streams of dual-stream networks, facilitating the learning of spatial-temporal consistency across multi scales.  Since our MSTI-Net comprehends the normality of spatial-temporal data both individually and jointly, it excels in detecting appearance-only, motion-only, and joint anomalies.

The main contributions of our work can be summarized as follows:
\begin{itemize}
    \item To enhance the spatial-temporal consistency of dual streams across multiple scales and concurrently filter out surplus background data during their interaction, we propose a Multi-scale Spatial-Temporal Interaction Network (MSTI-Net) for unsupervised VAD.
    \item In order to concentrate more acutely on local moving objects, we construct a unique Attention-based Spatial-Temporal Fusion Module (ASTFM). Furthermore, we incorporate multi-ASTFM-based connections between two streams to facilitate multi-scale spatial-temporal integration. This integration is key to effectively learning and internalizing the spatial-temporal coherence.
    \item The experimental results on three benchmark datasets  prove the effectiveness  of MSTI-Net, achieving 96.8\%, 87.6\%, and 73.9\% AUCs on the UCSD Ped2, CUHK Avenue, and ShanghaiTech datasets, respectively.
\end{itemize}

\begin{figure*}[t]
  \centering
  \includegraphics[width=0.8\textwidth]{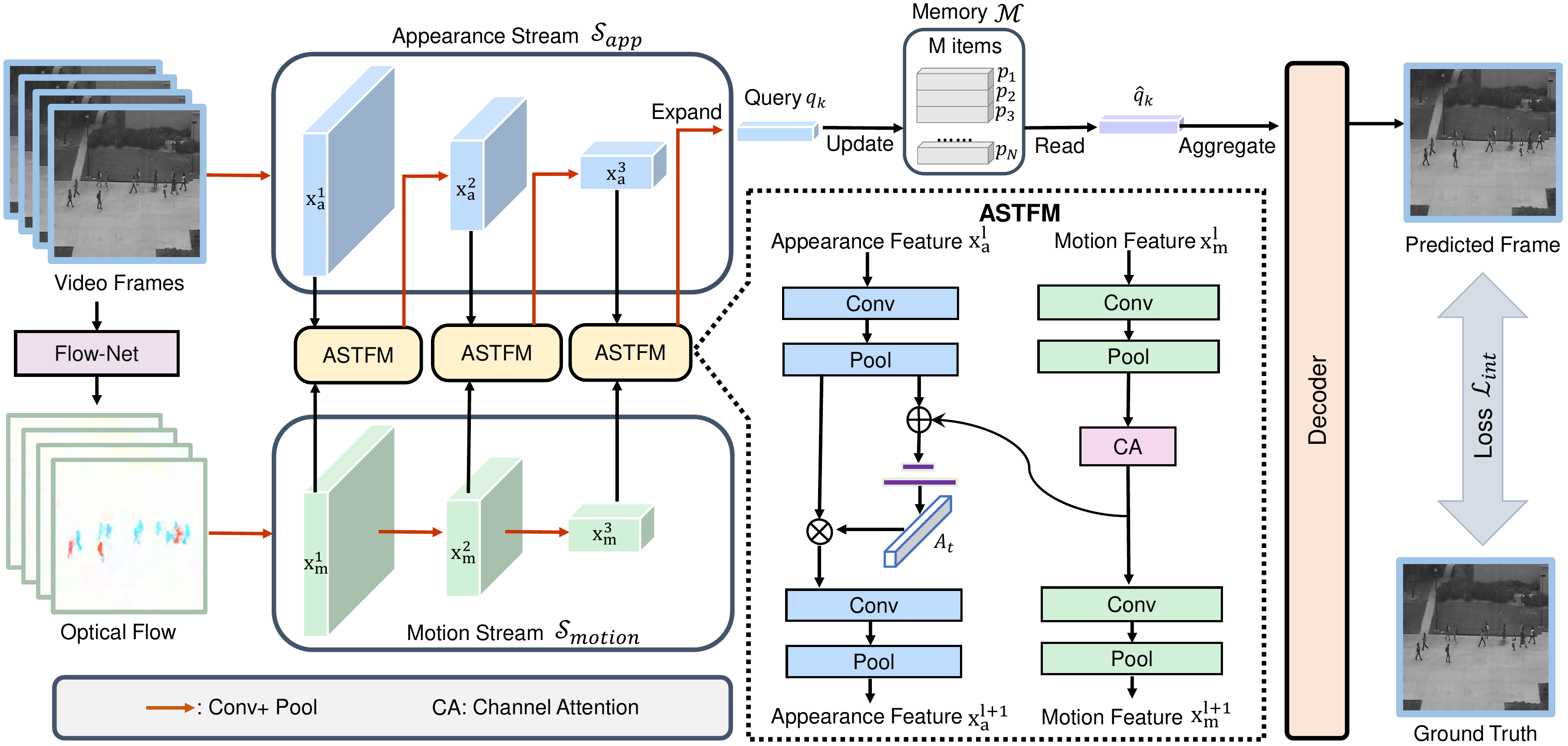}% 插入图片
  \caption{Overall framework of the proposed MSTI-Net. The appearance stream (upper branch) focuses on learning the visual patterns (e.g., shapes, colors, and textures) of regular events. The motion stream (lower branch) is designed to capture the changes in visual patterns over time. We integrate both the appearance and motion streams via multi Attention-based Spatial and Temporal Fusion Module (ASTFM)-based pathways to learn spatial-temporal coherence.} % 添加标题
  \label{fig1}% 添加标签以便引用
  \vspace{-10pt}
\end{figure*}

\section{Methodology}  % 标题用\section
\subsection{Overall Architecture} % 二级标题用\subsection

As shown in Fig. \ref{fig1}, our proposed MSTI-Net comprises five key components: the appearance stream,  the motion stream, the Attention-based Spatial-Temporal Fusion Module (ASTFM), the memory, and the decoder. For input data, a series of frames and their corresponding optical flow (calculated by well-trained Flownet\cite{flownet}) are fed into the appearance stream and motion stream respectively. These parallel streams are specifically designed to learn appearance patterns in the spatial dimension and motion patterns in the temporal dimension. To enhance this learning, both streams are structured with three distinct encoding levels.

Within each level, an ASTFM-based connection is injected from the motion stream to the appearance stream. The feature map of the motion stream is duplicated and then fused with the down-sampling appearance feature map through ASTFM, yielding the fused spatial-temporal features. These fused features subsequently pass through a convolution block and a down-sampling layer in sequence, facilitating the creation of the next-level feature map.

Post the third encoding level, the motion information is embedded into the appearance stream. The resultant output from the appearance stream is directed toward the memory module, which records the spatial-temporal consistency of regular events. Lastly, the output from the memory module is supplied to the decoder to predict future frames. Our MSTI-Net is trained to learn the normal patterns of appearance, motion, and joint characteristics. Consequently, during the testing phase, anomalies that are based solely on appearance, motion, or a combination of both, would generate large prediction errors. These errors can then be leveraged as a reference for computing regularity scores.

\subsection{Attention-based Spatial-Temporal Fusion Module} % 二级标题用\subsection

Due to the presence of redundant background information in video frames and the significant semantic disparity between the two streams, the direct transfer of signals (e.g., concatenating) from one network stream to another could inadvertently impair performance. To this end, we design an innovative fusion module known as the Attention-based Spatial-Temporal Fusion Module (ASTFM). The module focuses on fusing RGB and optical stream information using an attention-based mechanism, which is performed indirectly by multiplying the attention weight with the spatial features. This approach ensures a more fine-grained integration of data from both streams, thus preventing performance degradation. In our implementation, ASTFM is deployed on spatial-temporal connections across multiple scales.

\begin{figure}[t]
  \includegraphics[width=\columnwidth]{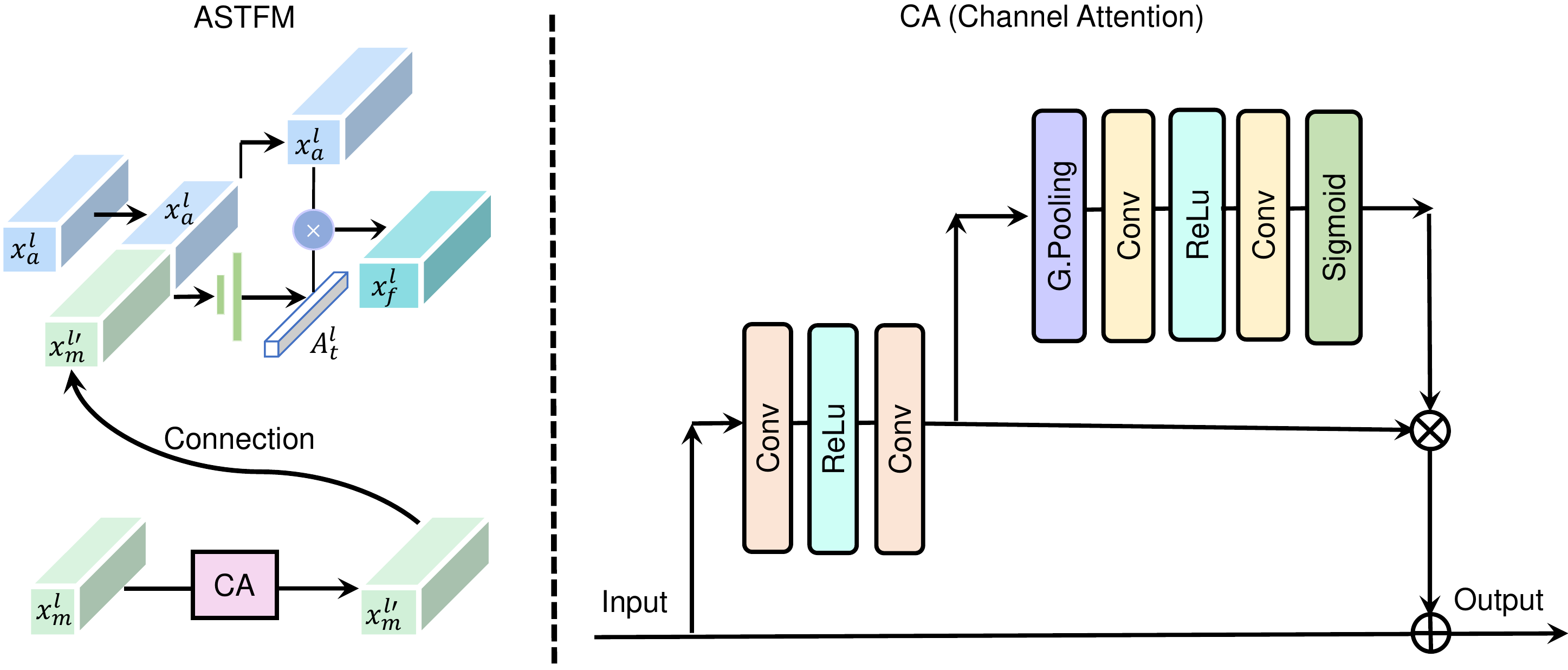}% 插入图片
  \caption{The structure of ASTFM (left) and Channel Attention (right).}% 添加标题
  \label{fig2}% 添加标签以便引用
  %\vspace{-10pt}
\end{figure}

As illustrated in Fig. \ref{fig2}, we first exploit channel attention\cite{le2023attention} to extract fine-grained motion information from the $l$-th layer motion feature acutely. Given the $l$-th layer input motion feature ${x}_{m}^{l}\in \mathbb{R}^{C \times H \times W}$, the computation of the channel attention block proceeds as follows:
\begin{equation}
\setlength{\abovedisplayskip}{3pt}
\setlength{\abovedisplayskip}{3pt}
    {{x}_{m}^{l}}^{\prime}={x}_{m}^{l}\oplus\left(g({x}_{m}^{l})\otimes \sigma(g({x}_{gp}))\right),
\end{equation}
where ${x}_{m}^{l}$ and ${{x}_{m}^{l}}^{\prime}$ represents input and output feature, ${x}_{gp}$ is the feature post global pooling. The symbol $\sigma(\cdot)$ corresponds to the sigmoid function, and $g(x)$ can be obtained by:
\begin{equation}
   g(x)=\bm{j}_{2}\text{R}\left(\bm{j}_{1} x\right),
\end{equation}
where $\text{R}(\cdot)$ denotes the ReLU activation function. The symbols $\bm{j}_{1}$ and $\bm{j}_{2}$ stand for the weight sets associated with the first and second convolutional layers, respectively.

Subsequently, we integrate the channel attention-enhanced $l$-th layer motion feature ${{x}_{m}^{l}}^{\prime}$ and corresponding-resolution appearance feature ${x}_{a}^{l}$. Specifically, ${{x}_{m}^{l}}^{\prime}$ and ${x}_{a}^{l}$ are initially concatenated, and the resulting concatenated feature $x_{cat}^{l}$ then passes through fully connected layers to generalize spatial-temporal attention weight $A_{t}^{l}$ for appearance feature $x_{a}^{l}$. The appearance features are refined by multiplying them with the attention weight $A_{t}^{l}$. The calculation of ASTFM is represented as follows:

\begin{equation}
{x}_{f}^{l} ={x}_{a}^{l}\otimes\mathcal{F}\left(({x}_{a}^{l}\oplus{{x}_{m}^{l}}^{\prime}); \boldsymbol{\mathcal{W}_{l}}\right) ,
\end{equation}
where ${x}_{f}^{l}$ signifies the output fused features, and $\mathcal{F}$ is a nonlinear mapping represented by convolutional filter weights $\boldsymbol{\mathcal{W}_{l}}$. Then, ${x}_{f}^{l}$ sequentially traverses a convolution block and a downsampling layer of the appearance stream.

\subsection{Memory}
In order to store the various prototypical patterns of the normal spatial-temporal consistency, we set the memory module following the appearance stream. The memory is configured as a two-dimensional matrix $\mathcal{M} = \left\{\boldsymbol{p}_{1}, \boldsymbol{p}_{2}, \cdots, \boldsymbol{p}_{N}\right\} \in \mathbb{R}^{N \times C}$, composed of N memory items, each of C dimensions. The memory module is responsible for the reading and updating of these items through the following processes.

Initially, the encoded prototypical feature $\bm{y}$ is read in memory and expanded into the query matrix $\boldsymbol{Q}=\left\{\boldsymbol{q}_{1}, \boldsymbol{q}_{2}, \cdots, \boldsymbol{q}_{N^{\prime}}\right\} \in \mathbb{R}^{N^{\prime} \times C}$. Subsequently, these queries are reintegrated by normal similar items in the following manner:

\begin{equation}
\hat{\bm{y}}=\bm{w}^{\top} \mathcal{M} ,
\end{equation}
where $\hat{\bm{y}}$ denotes the reintegrated prototypical features, and $\bm{w}^{\top}$ is a normalized weight vector computed by cosine similarity between every query $\boldsymbol{q}_{k} \left(\boldsymbol{k}=1,2, \ldots, \boldsymbol{N}^{\prime}\right)$ from Q and the full set of memory items $\boldsymbol{p}_{i} \left(\boldsymbol{i}=1,2, \ldots, \boldsymbol{N}\right)$. $\mathcal{M}$ represents all items in the memory module. Then, the item $\boldsymbol{p}_{i}$ in memory undergoes an update to capture robust patterns of normal events by the following computation:
\begin{eqnarray}
\hat{\boldsymbol{p}}_{i} = L_{2}\left(\boldsymbol{p}_{i}+\boldsymbol{v}_{i} \boldsymbol{Q}\right) ,
\end{eqnarray}
where $L_{2}$ represents the $l_{2}$ norm, $\bm{Q}=\left\{\boldsymbol{q}_{1}, \boldsymbol{q}_{2}, \cdots, \boldsymbol{q}_{N^{\prime}}\right\} \in \mathbb{R}^{N^{\prime} \times C}$is the query matrix, and $\bm{v}_{i}\in \mathbb{R}^{N \times C}$denotes the weight vector. The vector is determined by calculating the cosine similarity of $\boldsymbol{p}_{i} $ to all $\boldsymbol{q}_{k} \left(\boldsymbol{k}=1,2, \ldots, \boldsymbol{N}^{\prime}\right)$ and applying the Softmax operation.

\subsection{Training Loss}
Since the aim of the learning model is to produce subsequent frame $\bm{\hat{I}}_{t+1}$ that closely resembles the ground truth  $\bm{I}_{t+1}$ during the training phase, we calculate the intensity penalty $\mathcal{L}_{int}$ of each frame to train the model. The intensity penalty $\mathcal{L}_{int}$ based on ${L}_2$ norm  guarantees the similarity across all pixels within the RGB dimensional space. We construct $\mathcal{L}_{int}$ as follows:

\begin{equation}
\mathcal{L}_{int}\left(\hat{\bm{I}}_{t}, \bm{I}_{t}\right)=\left\|\hat{\bm{I}}_{t}-\bm{I}_{t}\right\|_{2}^{2}.    
\end{equation}

Besides, to  maintain sparsity and diversity within the memory items, we employ two distinct losses: the separateness loss $\mathcal{L}_{separate}$, and the compactness loss $\mathcal{L}_{compact}$. The value $\delta$  symbolizes a positive digit that presides over the acceptance level of dissimilarities. The definitions are as follows:
\begin{equation}
\mathcal{L}_{separate}=  \sum_{i=1}^{N}-w_{i} \cdot \log \left(w_{i}\right),
\end{equation}

\begin{equation}
\mathcal{L}_{compact}=  \max \left(\left(\left|\frac{\hat{\bm{y}}^{\top} \bm{y}}{\|\hat{\bm{y}}\|\|\bm{y}\|}\right|-\delta\right), 0\right).
\end{equation}

The final training loss $\mathcal{L}$ of our proposed MSTI-Net can be represented as follow, where $\lambda_{\mathrm{i}}$,  $\lambda_{\mathrm{s}}$ and $\lambda_{\mathrm{c}}$ denote the trade-off hyper-parameters:
\begin{equation}
\mathcal{L}=\lambda_{\mathrm{i}}\mathcal{L}_{int}+\lambda_{\mathrm{s}}\mathcal{L}_{separate}+\lambda_{\mathrm{c}}\mathcal{L}_{compact}.
\end{equation}

\subsection{Regularity Score}
During the testing phase, the anomaly detection ability of our MSTI-Net is  evaluated by the regularity score corresponding to each frame. Following the prior research\cite{ffp}, we utilize the Peak Signal-to-Noise Ratio (PSNR) between the estimated future frame $\bm{\hat{I}}_{t+1}$  and corresponding actual future frame $\bm{I}_{t+1}$. Concurrently, we also consider the distance between the query and its closest matching item in the memory set $\mathcal{M}$, to derive the regularity score ${R}_{t}$:
\begin{equation}
{R}_{t}=1-\tau(1-f(\operatorname{P}(\bm{\hat{I}}_{t+1}, \bm{I}_{t+1})))-(1-\tau) f\left(D\left(\boldsymbol{q}, \boldsymbol{p}\right)\right),   
\end{equation}
where $\tau$ denotes the balanced hyperparameter,  ${P}(\cdot,\cdot)$ represents PSNR operation, and ${f}(\cdot)$ is a normalization function. Moreover, ${D}(\cdot,\cdot)$ indicates the $\mathrm{L}_{2}$ distance between all queries and closest matching item in $\mathcal{M}$.

\section{Experiments}
\subsection{Implementation}
The validity of our MSTI-Net is evaluated on three benchmark datasets: UCSD Ped2\cite{li2013anomaly}, CUHK Avenue\cite{lu2013abnormal}, and ShanghaiTech dataset\cite{luo2017revisit}. Following the prior work\cite{zhong2022bidirectional}, we employ the frame-level Area Under the Curve (AUC) as a key performance indicator to evaluate the performance of our proposed methodology. In our implementation, each video frame is resized to 256×256 dimensions and normalized to fit the range of [-1, 1]. Adam optimizer is employed for model optimization with an initial learning rate set at $2\times 10^{-4}$. The hyperparameters $\lambda_{i}$, $\lambda_{\mathrm{s}}$, $\lambda_{\mathrm{c}}$, $\delta$ and $\tau$ are empirically set to 0.8, $3\times 10^{-4}$, 0.001, 0.1 and 0.7, respectively.

\subsection{Comparison with Existing Methods}

\begin{table}[t]
    \centering
    \caption{Quantitative comparison results of frame-level AUC on the three benchmark datasets. The best-performing results are highlighted in bold for emphasis.}
    \label{tab1}
    \resizebox{.44\textwidth}{!}{
    \begin{tabular}{@{}cccccccc@{}}
    \toprule
    \text Type & \text Method & \text UCSD Ped2  & \text CUHK Avenue   & \text ShanghaiTech          \\ \midrule

    \multirow{4}{*}{(1)} 
    &\text ConvLSTM-AE \cite{luo2017remembering}     & 88.1\% & 77.0\%&  -  \\
    &\text Multispace\cite{zhang2020normality} & 95.4\% & 86.8\% &  73.6\%   \\ 
    &\text IPR\cite{tang2020integrating}     & 96.2\% & 83.7\% &  71.5\%   \\
    &\text DAF\cite{smeureanu2017deep}     & - & 84.6\% &  - \\  \midrule
    
    \multirow{4}{*}{(2)} 
    &\text FFP\cite{ffp}     & 95.4\% & 85.1\% &  72.8\% \\ 
    &\text CDDA\cite{chang2020clustering}   & 96.5\% & 86.0\% &   73.3\%   \\
    &\text AMC \cite{nguyen2019anomaly}    & 96.2\% & 86.9\% &  -\\ \midrule
    
\multirow{3}{*}{(3)} 
    &\text SIG-Net\cite{fang2020anomaly} & 96.2\% & 86.8\% &  -\\
    &\text DDGA-Net\cite{dong2020dual} & 95.6\% & 84.9\% &73.7\% \\
    &\text AMMC-Net\cite{AMMC}     & 96.6\% & 86.6\% &  73.7\%  \\
    &\text STD\cite{chang2022video}     & 96.7\% &87.1\% &73.7\%\\
    \midrule

    \multirow{1}{*}{(4)} 
    &\text Ours      & \textbf{96.8\%} & \textbf{87.6\%} &  \textbf{73.9\%}  \\ 
    \bottomrule
    \end{tabular}}
    % \vspace{-5pt}
\end{table}

\begin{table}[t]
    \centering
    \caption{Outcomes of our ablation study. We detail the frame-level AUC for every model variation on the ShanghaiTech dataset.}
    \label{tab2}
    \resizebox{.44\textwidth}{!}{
    \begin{tabular}{@{}cccccccc@{}}
    \toprule
    Model & $\mathcal{S}_{app}$   & $\mathcal{S}_{motion}$   & $\mathcal{M}$      & $\mathcal{STC}$   & $\mathcal{ASTFM}$  & AUC (\%)      \\ \midrule
    A     &\Checkmark& &\Checkmark& & &73.1\%   \\
    B     &\Checkmark&\Checkmark&\Checkmark& &  & 73.3\%   \\
    C    &\Checkmark&\Checkmark& \Checkmark&\Checkmark& & 72.6\%   \\ 
    D  &\Checkmark&\Checkmark& & \Checkmark&\Checkmark& 72.8\%   \\
    E
    &\Checkmark &\Checkmark & \Checkmark&\Checkmark & \Checkmark& \textbf{73.9\%} \\ 
\bottomrule
    \end{tabular}}
    %\vspace{-5pt}
\end{table}

\begin{figure}[t]
  \includegraphics[width=\columnwidth]{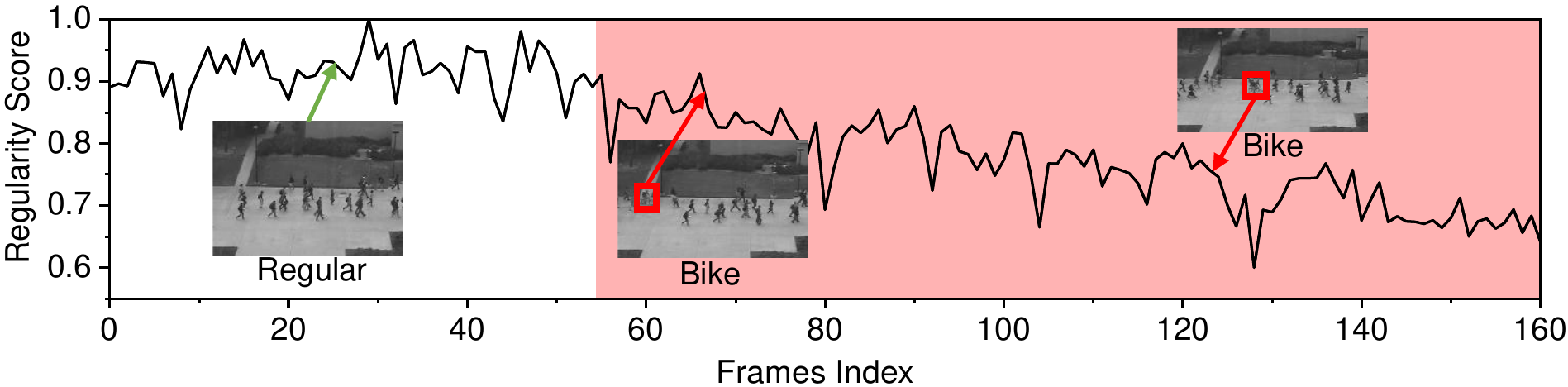}% 插入图片
  \caption{Regularity score of the test video derived in the UCSD Ped2 dataset. The period of anomalous activity (e.g., bike riding) is highlighted in pink.}% 添加标题
  \label{fig3}% 添加标签以便引用
\end{figure}

\begin{figure}[t]
  \includegraphics[width=\columnwidth]{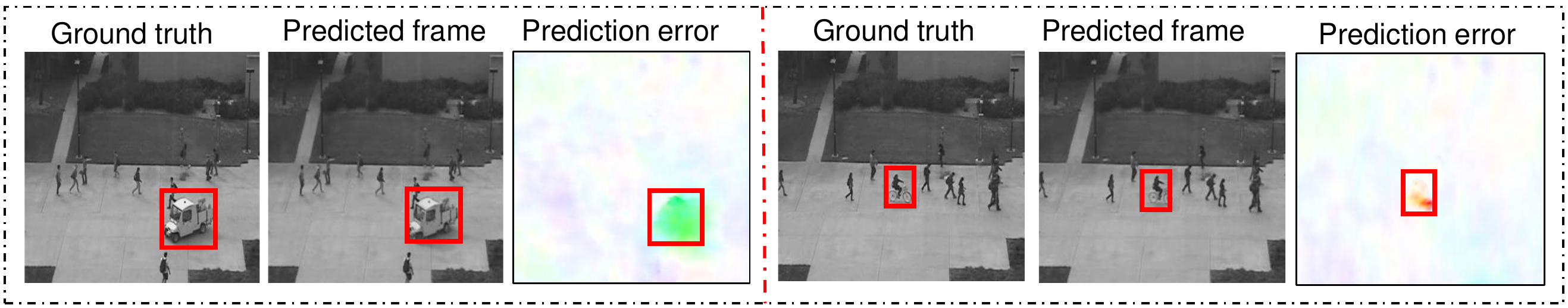}% 插入图片
  \caption{Visual display of predicted frames and associated prediction errors. The red bounding boxes are used to mark the positions of abnormal objects.}% 添加标题
  \label{fig4}% 添加标签以便引用
  % \vspace{-15pt}
\end{figure}

Table \ref{tab1} shows the frame-level AUC results of the proposed MSTI-Net alongside several previous methods on three benchmark datasets. A higher AUC score signifies superior performance. The compared methods can be grouped into three categories:  (1) the single-stream methods\cite{luo2017remembering,zhang2020normality,tang2020integrating,smeureanu2017deep}, (2) dual-stream networks\cite{chang2022video,ffp,chang2020clustering,nguyen2019anomaly}, and (3) interaction networks\cite{fang2020anomaly,dong2020dual,AMMC,chang2022video}. Typically, the performance of the dual-stream networks (2) exceeds that of the single-stream methods (1), suggesting that learning normality in a spatial-temporal manner can more effectively model the patterns of normal events. Furthermore, the performance of  our MSTI-Net (4) is better than both the traditional dual-stream network (2) and  other interaction networks (3). This demonstrates the importance and effectiveness of learning spatial-temporal consistency from multi scales. In summary, compared to the techniques outlined in Table I, our MSTI-Net achieves state-of-the-art performance on the UCSD Ped2, CUHK Avenue, and ShanghaiTech datasets.

\subsection{Ablation Study}

Table \ref{tab2} displays the ablation study results for the ShanghaiTech dataset. We conduct ablation experiments for ASTFM, spatial-temporal connections $\mathcal{STC}$, appearance stream $\mathcal{S}_{app}$, motion stream $\mathcal{S}_{motion}$, and Memory $\mathcal{M}$. As shown in table \ref{tab2}, the dual-stream interaction structure with ASTFM-based connections (model E) shows an AUC improvement of 0.8\% and 0.6\% compared to the single-stream structure (model A) and the dual-stream separate structure without interactions (model B), respectively. This indicates the efficacy of the spatial-temporal interaction within the dual-stream network. Furthermore, compared to model C (which uses traditional fusion), model E (employing ASTFM-based interaction) exhibits a 0.3\% AUC improvement, thereby validating the effectiveness of our novel attention-based fusion approach. Lastly, the comparison between model D (without memory) and model E (with memory) affirms the utility of the memory module.

\subsection{Visualization Analysis}
To further demonstrate the effectiveness of our method on the VAD task, we conduct a visualization analysis, the results are shown in Fig. \ref{fig3}, and Fig. \ref{fig4}. 
In Fig. \ref{fig3}, we visualize regularity scores output by MSTI-Net on testing videos from UCSD Ped2. A high regularity score signifies a superior prediction outcome. Noticeably, the regularity score experiences a sharp decline when an anomaly occurs (denoted by the pink region in Fig. \ref{fig3}). Fig. \ref{fig4} depicts the prediction error for two anomalous frames by comparing the ground truth with the corresponding predicted frame. It can be observed that objects moving anomalously (e.g., bicycles, cars) have higher prediction errors. Therefore, this visualization analysis further indicates the ability of our MSTI-Net in detecting anomalies in video sequences.

\section{Conclusion}
This study primarily addresses the problem that prevalent dual-stream networks limit spatial-temporal interaction and inadvertently integrate background noise into the fusion process. For this purpose, we design the MSTI-Net to model the spatial-temporal consistency information at multi scales and design a novel Attention-based Spatial-Temporal Fusion Module (ASTFM) to prioritize the detection and analysis of moving objects in the scene. The experimental results on three benchmark datasets illustrate that MSTI-Net surpasses existing state-of-the-art solutions. We further validate the effectiveness of MSTI-Net's key components through comprehensive ablation studies and visualization analyses. Our future endeavors will be centered on unearthing enhanced techniques to understand spatial-temporal consistency within the Video Anomaly Detection realm.

\balance

\bibliographystyle{IEEEbib}
\bibliography{refs}

\end{document}